\documentclass[twoside]{article}

\usepackage[round]{natbib}

\usepackage{pifont}
\usepackage{stfloats}
\usepackage{fullpage}
\usepackage[utf8]{inputenc} 
\usepackage[T1]{fontenc}    
\usepackage{hyperref}
\usepackage{booktabs}       
\usepackage{amsfonts,amsmath,amsthm,amssymb}       
\usepackage{nicefrac}       
\usepackage{microtype}      
\usepackage{graphicx}

\usepackage{mathtools}
\usepackage{mathrsfs}
\usepackage{amsthm}
\newcommand{\x}{\mathbf{x}}
\newcommand{\z}{\mathbf{z}}

\renewcommand{\u}{\mathbf{u}}
\newcommand{\y}{\mathbf{y}}

\renewcommand{\t}{\mathbf{t}}

\newcommand{\f}{\mathbf{f}}

\newcommand{\Y}{\mathbf{Y}}
\newcommand{\btau}{\boldsymbol{\tau}}

\newcommand{\X}{\mathbf{X}}

\newcommand{\bt}{{\boldsymbol{\theta}}}

\newcommand{\Z}{\mathbf{Z}}
\newcommand{\R}{\mathbb{R}}

\newcommand{\N}{\mathcal{N}}
\newcommand{\GP}{\mathcal{GP}}

\newcommand{\bmu}{\boldsymbol{\mu}}

\newcommand{\bS}{{\boldsymbol{\Sigma}}}

\newcommand{\bL}{{\boldsymbol{\Lambda}}}

\newcommand{\bomega}{\boldsymbol{\omega}}

\newcommand{\norm}[1]{\left\lVert#1\right\rVert}

\makeatletter \renewcommand\d[1]{\ensuremath{%
  \;\mathrm{d}#1\@ifnextchar\d{\!}{}}}
\makeatother

\theoremstyle{definition}

\title{Learning spectrograms with convolutional spectral kernels}

\author{ Zheyang Shen \qquad Markus Heinonen \qquad Samuel Kaski  \\ Helsinki Institute for Information Technology, HIIT \\ Department of Computer Science, Aalto University}

\begin{document}
\twocolumn[\maketitle]
\begin{abstract}
  We introduce the \emph{convolutional spectral kernel} (CSK), a novel family of non-stationary, nonparametric covariance kernels for Gaussian process (GP) models, derived from the convolution between two imaginary radial basis functions. We present a principled framework to interpret CSK, as well as other deep probabilistic models, using approximated Fourier transform, yielding a concise representation of input-frequency \emph{spectrogram}. Observing through the lens of the spectrogram, we provide insight on the interpretability of deep models. We then infer the functional hyperparameters using scalable variational and MCMC methods. On small- and medium-sized spatiotemporal datasets, we demonstrate improved generalization of GP models when equipped with CSK, and their capability to extract non-stationary periodic patterns.
\end{abstract}
\section{Introduction}
Gaussian processes (GP), as rich distributions over functions, are a cornerstone of a wide array of probabilistic modeling paradigms, thanks largely to their tractability, flexibility, robustness to overfitting and principled quantification of uncertainty \citep{rasmussen}. At the helm of every GP model lies the \emph{covariance kernel}, a function depicting its covariance structure and encoding prior knowledge.\par
Despite their equivalence to infinitely wide neural networks \citep{williams1997computing}, GP models seldom exhibit the generalization of the former due to the innate rigidity of the commonly used squared exponential (SE) kernel, rendering them insufficient for genuine pattern recognition. While myriad studies \citep{paciorek2004nonstationary, alvarez2009latent, wilson2013gaussian, pmlr-v28-duvenaud13, tobar2015learning, wilson2016deep, remes2017non, tobar2018bayesian,shen2019harmonizable} have sought more expressive kernel choices, their efforts fall notably short on (i) flexibility, (ii) interpretability or (iii) scalability, all of which are essential to large-scale statistical analysis. In this work, we propose and analyze a novel kernel family satisfying the above three properties.\par
We propose the \emph{convolutional spectral kernel} (CSK), a novel kernel family with both spatially varying lengthscales and frequencies, derived from the convolution of two complex radial basis functions. We demonstrate that CSK possesses superior flexibility, unifying the monotonic non-stationary quadratic (NSQ) kernel \citep{paciorek2004nonstationary} and the stationary spectral mixture (SM) kernel \citep{wilson2013gaussian}.\par
We introduce the notion of the \emph{spectrogram} as a new, principled framework to interpret nonparametric kernels. The spectrogram is a joint distribution of input and frequency, conveniently displaying local covariance patterns. Our analysis shows that CSK retains an unbiased description of the instantaneous frequency, as opposed to the similar generalized spectral mixture (GSM) kernel \citep{remes2017non}. Meanwhile, our analysis sheds light on previously un-interpretable state-of-the-art deep probabilistic models, namely deep GPs \citep{damianou2013deep, salimbeni2017doubly, havasi2018inference} and the deep kernel learning \citep{wilson2016deep}, and justifies the adoption of certain heuristics in the said models.\par
We introduce scalable inference schemes for GP models equipped with CSK, which combine sparse GPs \citep{snelson2006sparse, titsias2009variational, hensman2017variational}, stochastic gradient Hamiltonian Monte Carlo \citep{Neal93probabilisticinference, chen2014stochastic}, and the recently proposed moving window MCEM \citep{havasi2018inference}. Our method can be extended to covariance function deep GPs \citep{dunlop2018deep}, a hierarchical generalization of our current model.\par
Lastly, we provide empirical evidence from both synthetic and real-world spatiotempral datasets to support our theoretical claims. Our experiments visually and numerically demonstrate interpretable pattern extraction, along with superior predictive performance of the CSK-GP model.\par
\section{Convolutional spectral kernel (CSK)}
\begin{figure*}[t]
    \centering
    \includegraphics[trim={3cm 2cm 2cm  2cm},clip,width=\textwidth]{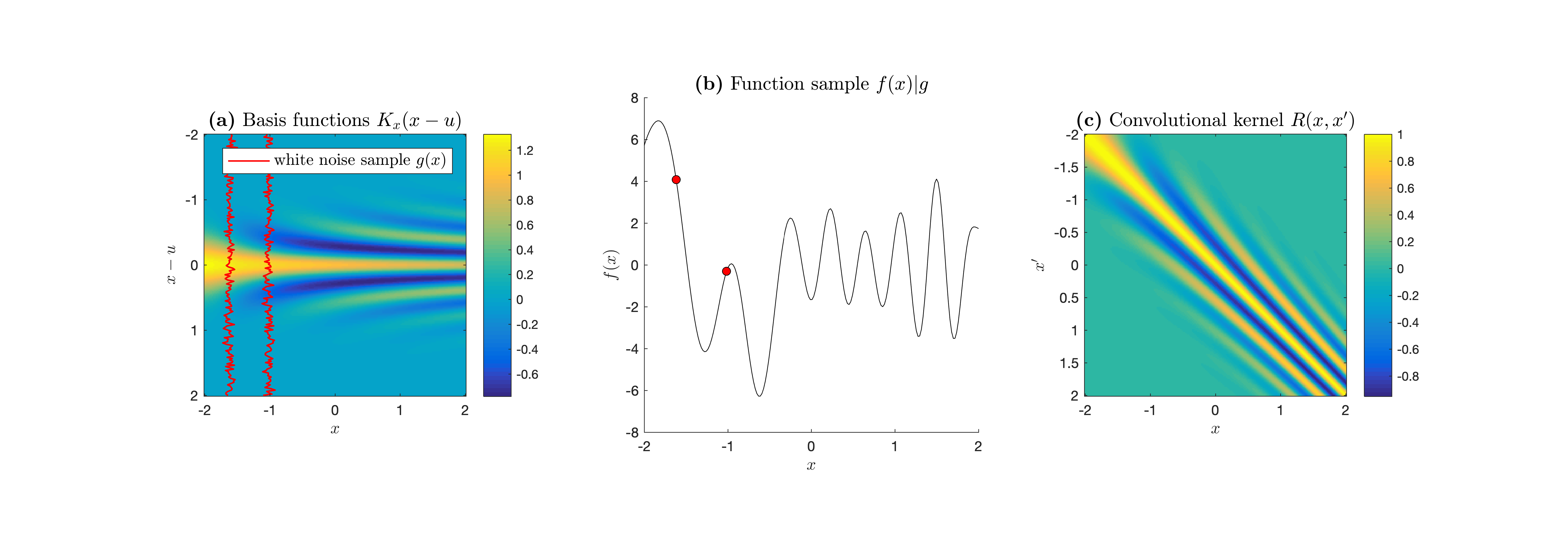}
    \caption{Visualization of the \textbf{(a)} basis function and convolution with a white noise process; \textbf{(b)} a sample from the resulting GP, where the highlighted points are computed from a convolution between the moving white noise process, denoted by the jagged lines in \textbf{(a)}; \textbf{(c)} the kernel matrix of the resulting CSK.}
    \label{fig:conv_vis}
\end{figure*}
In this section, we derive the \emph{convolutional spectral kernel}, a non-stationary, nonparametric kernel, interpretable through its local lengthscale, frequency and variance functions. Throughout the discussion of this paper, we assume a simple regression task: the objective is to infer a scalar function $f(\x) \in \R$ with $D$-dimensional inputs $\x \in \R^D$, with a finite supply of $N$ observed data points as a data matrix $\X \in \mathbb{R}^{N\times D}$, and a set of noisy observations $\y \in \mathbb{R}^N$. We assume the function $f$ is a realization of some underlying zero-mean Gaussian process, with homoskedastic observation noise of precision $\beta$,
\begin{align}
    f(\x) &\sim \GP(0, k(\x, \x')), \\
    y &= f(\x) + \epsilon, \qquad \epsilon \sim \N(0, \beta^{-1}) \label{eq:obs}.
\end{align}
Our construction of CSK is inspired by the construction of non-stationary kernels with spatially-varying lengthscales \citep{gibbs1998bayesian, higdon1998,paciorek2004nonstationary}. We propose a novel feature map $K_{\x}(\u) \in \mathbb{C}$ of complex-valued radial bases:
\begin{align}
\small
    K_{\x_i}(\u) &= e^{-\frac{\norm{\bS_i^{-1/2}(\u-\x_i)}^2}{2}}\exp(\imath\langle\bS_i^{-1}\bmu_i,\u-\x_i\rangle) \notag \\
    &\propto \N\left(\u|\x_i+\imath\bmu_i, \bS_i\right). \label{eq:feature_map}
\end{align}
Here we abuse the notation of a multivariate normal density, where $\imath$ denotes the imaginary unit, $\u \in \R^D$ denotes a point in input space, and the $\bmu_i := \bmu(\x_i) \in \R^{D}$, $\bS_i := \bS(\x_i) \in \R_{\succeq \mathbf{0}}^{D \times D}$ are vector- and positive-semidefinite matrix-valued functions of $\x_i$ denoting the frequency and covariance parameters of the input space, from which we can construct the \emph{frequency} as an inverse product $\bS^{-1} \bmu$. Viewing GPs as continuously-indexed moving average processes, the feature map \eqref{eq:feature_map} denotes a potentially infinite window \citep{tobar2015learning}. We can henceforth represent a GP as a convolution between $K_{\x_i}$ and a white noise process $g(\x)\sim\GP(0, \delta_{\x=\x'})$ \citep{higdon1998}:
\begin{align}
    f(\x_i) = \int K_{\x_i}(\u) g(\u) \d{\u}.
\end{align}
The kernel of $f$ is the Hermitian inner product between $K_{\x_i}(\u)$ and $K_{\x_j}(\u)$, which is solved analytically:
\begin{align}
    k(\x_i, \x_j) &= \int K_{\x_i}(\u) \overline{K_{\x_j}(\u)}\d{\u} \notag\\
    &\propto \N\left(\x_i-\x_j|\imath(\bmu_i+\bmu_j), \bS_i+\bS_j\right).
    \label{eq:csk_orig}
\end{align}
\begin{figure*}[t]
    \centering
    \includegraphics[width=\textwidth]{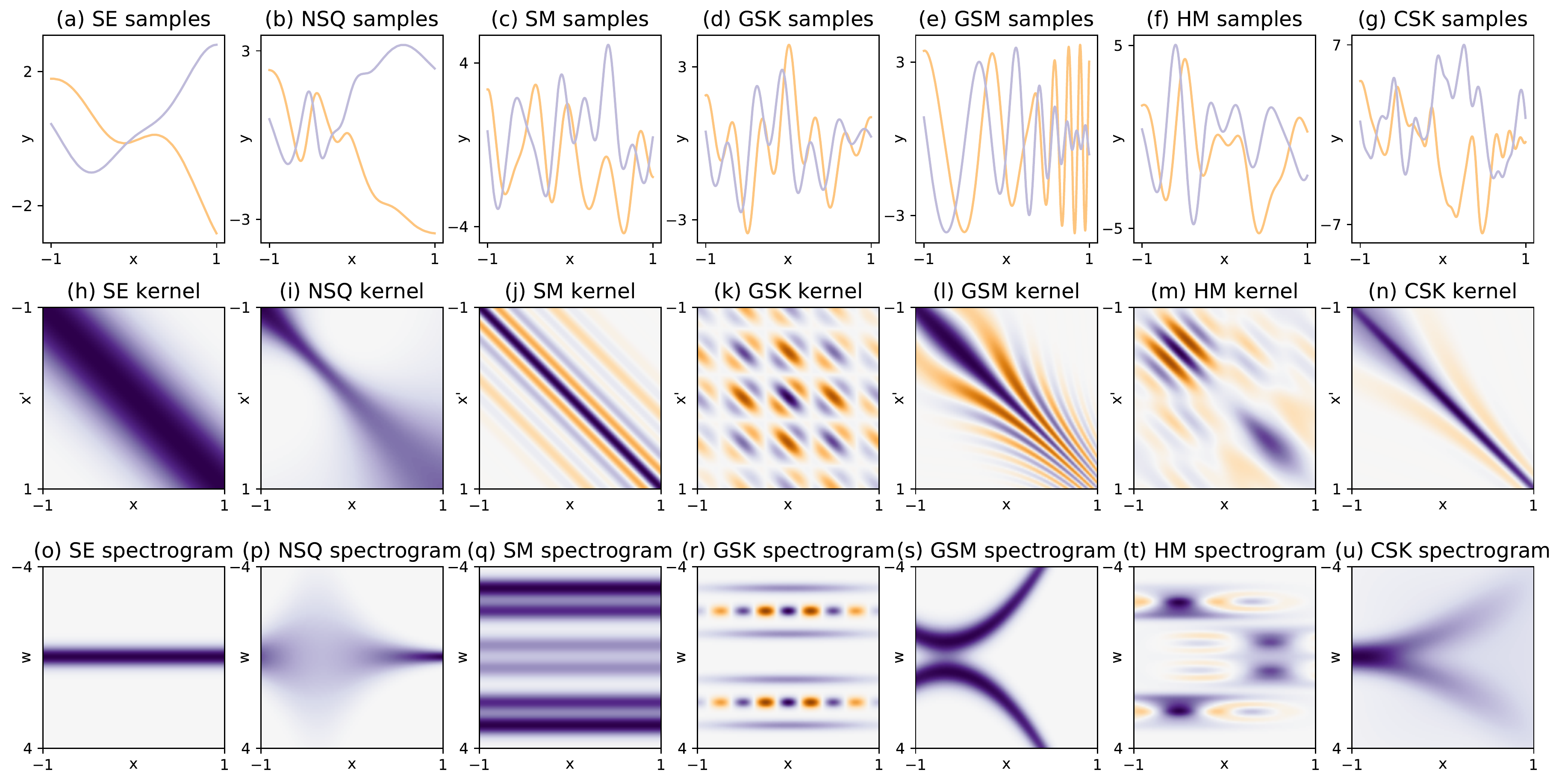}
    \caption{Overview of some kernels used in GP models, with sample paths (first row), kernel matrices (second row), and spectrogram (third row). Examples include SE \citep{rasmussen}, NSQ \citep{paciorek2004nonstationary}, SM \citep{wilson2013gaussian}, GSK \citep{samo2017advances}, GSM \citep{remes2017non}, HM \citep{shen2019harmonizable} and CSK (current work).}
    \label{fig:kernels}
\end{figure*}
Here $\overline{K_{\x_i}(\cdot)}$ denotes the complex conjugate. The solution to this integral is detailed in Section 1 of the appendix. We obtain a non-stationary correlation function after normalization:
\begin{align}
    R(\x_i, \x_j) &= \frac{\mathrm{Re}\left(k(\x_i, \x_j)\right)}{\sqrt{k(\x_i, \x_i)k(\x_j, \x_j)}} \notag \\
    &= \sigma_{ij}\  e^{-\frac{Q_{ij}+S_{ij}}{2}}\cos\langle\bomega_{ij},\x_i-\x_j\rangle, \label{eq:csk_norm}\\
    \sigma_{ij} &= \frac{|\bS_i|^{1/4}|\bS_j|^{1/4}}{\left|(\bS_i+\bS_j)/2\right|^{1/2}},\\
    Q_{ij} &= \norm{\left(\bS_i+\bS_j\right)^{-1/2}(\x_i-\x_j)}^2, \label{eq:qij}\\
    \bomega_{ij} &= \left(\bS_i+\bS_j\right)^{-1}(\bmu_i+\bmu_j),\label{eq:omegaij}\\
    S_{ij} &= \norm{\left(\bS_i^{-1}+\bS_j^{-1}\right)^{-1/2}(\bomega_{ii}-\bomega_{jj})}^2. \label{eq:sij}
\end{align}
We can see from the cosine term in \eqref{eq:csk_norm} that CSK computes \emph{pairwise frequencies} $\bomega_{ij}$ for each pair of data points, and the exponential terms include squared Mahalanobis distances between $\x_i$ and $\x_j$ \eqref{eq:qij}, and between local frequencies $\bomega_{ii}$ and $\bomega_{jj}$ \eqref{eq:sij}.\par
CSK unifies two generalizations of the SE kernel. \citet{paciorek2004nonstationary} generalize the SE kernel by allowing lengthscales to spatially vary, which is a special case of CSK when $\bmu_i\equiv\mathbf{0}$. Formally, 
\begin{align}
    k_{NS}(\x_i, \x_j) &= \sigma_{ij} e^{-\frac{Q_{ij}}{2}} \propto \N\left(\x_i|\x_j, \bS_i+\bS_j\right). \label{eq:nsq}
\end{align}
The spectral mixture kernel \citep{wilson2013gaussian} generalizes the SE kernel by allowing for non-zero frequency mean, which is a special case of CSK when functions $\bmu_i$ and $\bS_i$ are kept constant:
\begin{align}
    k_{SM}(\x_i, \x_j) &= k_{SE}(\x_i, \x_j)\cos\langle\bmu, \x_i-\x_j\rangle. \label{eq:sm}
\end{align}
CSK as a correlation function identifies one frequency component in the underlying data, while the data might exhibit behaviors such as multiple frequencies and spatially varying variances. Such behaviors can be accounted for by stacking multiple CSKs multiplied by a standard deviation function $\sigma_p(\cdot) \in \mathbb{R}_{\geq 0}$:
\begin{align}
    k_{\text{CS}}(\x_i, \x_j) &= \sum_{p=1}^P \sigma_p(\x_i)\sigma_p(\x_j)R_p(\x_i, \x_j). \label{eq:csk_final}
\end{align}
\hspace{-0.11cm}CSK is defined through the component functions $\sigma_p(\cdot)$, $\bS_p(\cdot)$, and $\bmu_p(\cdot)$. We denote the vector of functional parameters as $\bt$, and each functional parameter $\theta_d(\x)$ has a warped GP \citep{snelson2004warped} prior:
\begin{align}
    \theta_d(\x_i) &= F_d(h_d(\x_i)), \label{eq:warped}\\
    h_d(\x_i) &\sim \GP(0, k_d(\x_i, \x_j)),
\end{align}
For simplification, we assume diagonal covariances: $\bS_p=\text{diag}(\ell^2_{p1}, \cdots \ell^2_{pD})$. The warping function $F_d$ ensures the CSK to be positive definite. 
\section{The spectrogram}
\begin{table*}[b]
    \centering
    \resizebox{\textwidth}{!}{
    \begin{tabular}{lcccr} \toprule
        Kernel & form & $\boldsymbol{\xi}_{\x}$ & $\bL_{\x}$ & reference\\
        \hline
        Spectral mixture & \eqref{eq:sm} & $\bmu$ & $\bS^{-1}$ & \citet{wilson2015human}\\
        Non-stationary quadratic & \eqref{eq:nsq} & $\mathbf{0}$ & $\frac{1}{2}\bS_{\x}^{-1}$ & \citet{paciorek2004nonstationary}\\
        Generalized spectral mixture & \eqref{eq:gsm} & $\bmu_{\x}+ \left(\mathcal{J}_{\x}\bmu_{\x}\right) \x$ & $\frac{1}{2}\bS_{\x}^{-1}$ & \citet{remes2017non}\\
        Convolutional spectral kernel & \eqref{eq:csk_norm} & $\bS_{\x}^{-1}\bmu_{\x}$ & $\frac{1}{2}\left(\bS_{\x}^{-1} + \mathcal{J}_{\x}\left(\bomega_{\x\x}\right)^\top\bS_i\mathcal{J}_{\x}\left(\bomega_{\x\x}\right)\right)$ & current work\\
        DGP-SE & \eqref{eq:dgp_kernel} & $\mathbf{0}$ & $\mathcal{J}_{\x}\f_{L-1}^\top \bS^{-1} \mathcal{J}_{\x}\f_{L-1}$ & \citet{damianou2013deep}\\
    \bottomrule
    \end{tabular}
    }
    \caption{Spectrogram function parameters for various non-stationary covariance function and compositional DGP.}
    \label{tab:spectrograms}
\end{table*}
This section coins the notion of \emph{spectrogram}, a joint input-frequency distribution, as a principled framework to interpret typical nonparametric kernels encountered in GP models, which typically lacks interpretability.\par
In signal processing and time-series analysis, the Wigner transform \citep{flandrin1998time} converts covariance functions into quasi-probability distributions between input and frequency via a Fourier transform:
\begin{align}
    W(\x, \bomega) &= \int_{\R^D} k\left(\x+\frac{\btau}{2}, \x-\frac{\btau}{2}\right)e^{-2\imath\pi\bomega^\top\btau}\text{d}\btau. \label{eq:wigner}
\end{align}
The Wigner distribution function (WDF) $W(\x, \bomega)$ is a generalized probability distribution that retains instantaneous spectral density on all inputs. Despite their potential in interpretation, few machine learning models \citep{shen2019harmonizable} apply the transform \eqref{eq:wigner} due to its intractability.\par
In this work, we consider all kernels taking the form of a generalized mixture of Gaussian characteristic functions, as indexed by $\mathscr{A}$: 
\begin{align}
    k(\x_i, \x_j) &= \sum_{a\in\mathscr{A}} \sigma_{ij}^{(a)}e^{-\frac{D_{ij}^{(a)}}{2}}\exp(\imath U_{ij}). \label{eq:sm_generalized}
\end{align}
Here the $\sigma$, $D$ and $U$ are real-valued functions of $\x_i$ and $\x_j$. A linearized approximation of the above functions gives an estimate of the kernel value, where the variables $\x$ and $\btau$ are separated:
\begin{align}
    k^{(a)}\left(\x+\frac{\btau}{2}, \x-\frac{\btau}{2}\right) & \approx \sigma_{\x\x}^{(a)} e^{-\frac{1}{2}\btau^\top\bL^{(a)}_{\x}\btau}\exp(\imath \langle\boldsymbol{\xi}_{\x}^{(a)}, \btau\rangle), \label{eq:sm_gen_est}\\
    \bL_{\x}^{(a)} &= \left.\mathcal{H}_{\t}\ D_{\x+\t/2, \x-\t/2} \right\rvert_{\t=\mathbf{0}}, \label{eq:approx2}\\
    \boldsymbol{\xi}_{\x}^{(a)} &= \left.\mathcal{J}_{\t}\ U_{\x+\t/2, \x-\t/2} \right\rvert_{\t=\mathbf{0}}. \label{eq:approx3}
\end{align}
$\mathcal{H_{\t}}$ and $\mathcal{J_{\t}}$ in \eqref{eq:approx2} and \eqref{eq:approx3}, respectively, denote the Hessian and Jacobian operators with respect to the variable $\t$. We can henceforth approximate the WDF with the kernel estimate \eqref{eq:sm_gen_est}:
\begin{align}
    \widehat{W}(\x, \bomega) &= \sum_{a\in\mathscr{A}} \sigma^{(a)}_{\x\x} \N\left(\bomega\left\vert \frac{\boldsymbol{\xi}_{\x}^{(a)}}{2\pi}, \frac{\bL_{\x}^{(a)}}{2\pi^2}\right.\right). \label{eq:spectrogram}
\end{align}
Our approximation is exact for both stationary and harmonizable spectral kernels \citep{wilson2013gaussian, shen2019harmonizable}. The rest of the section delineates the significance of our method in interpreting CSK and deep GP (DGP) models. The models in interest and their spectrograms are listed in Table \ref{tab:spectrograms}, and their derivation summarized in Section 2 of the appendix.
\subsection{Spectrograms of non-stationary kernels}
The spectrogram applies to GP models equipped with nonparametric kernels \citep{paciorek2004nonstationary, damianou2013deep, remes2017non}, as demonstrated in Table \ref{tab:spectrograms}. In particular, we investigate the notable similarity between CSK and generalized spectral mixture (GSM) \citep{remes2017non} kernel.\par

The GSM kernel is a nonparametric, quasi-periodic kernel with an intuitively defined spectrogram. The correlation of GSM is a parametrization of \eqref{eq:sm_generalized}
\begin{align}
    D_{ij} &= \norm{\left(\bS_i+\bS_j\right)^{-1/2}(\x_i-\x_j)}^2,\\
    U_{ij} &= \langle \bmu_i, \x_i \rangle - \langle \bmu_j, \x_j \rangle. \label{eq:gsm}
\end{align}
While it is tempting to equate the frequency mean $\boldsymbol{\xi}_{\x}$ \eqref{eq:approx3} with $\bmu_i$, our analysis yields contradictory evidence, with $\boldsymbol{\xi}_{\x_i} = \bmu_i+ \left(\mathcal{J}_{\x}\bmu\right)\vert_{\x=\x_i} \x_i$, rendering the GSM kernel inherently biased in instantaneous frequency, which leads to an erroneously defined intuitive spectrogram \citep{remes2017non}.\par
The CSK records unbiased frequencies in contrast, albeit suffers a bias in the lengthscales (as shown in Table \ref{tab:spectrograms}). We posit that it is essential that we adopt unbiased frequencies, so that the kernel extracts correct periodicities.
\subsection{Standard DGPs are equivalent to GPs with NSQ kernels}
\label{sec:dgp_equiv}
Through spectrogram analysis, we uncover an equivalence between two classes of DGPs, namely the ones constructed with compositions and the ones with nonparametric covariance functions \citep{dunlop2018deep}.\par
The compositional DGP \citep{damianou2013deep, salimbeni2017doubly, dunlop2018deep} generalizes standard GP with recursive functional composition:
\begin{align}
    f &= f_L \circ f_{L-1} \circ \cdots \circ f_0, \label{eq:dgp_def}\\
    f_l & \sim \GP(m_l, k_l),\ l=0, \cdots L.
\end{align}
The conditional kernel $k_l|f_{l-1}$ can be seen as one with varying lengthscales when $k_L$ is the default SE kernel:
\begin{align}
    k_l(x,x^\prime|f_{l-1}) &= w^2 e^{-\frac{(f_{l-1}(x) - f_{l-1}(x^\prime))^2}{2\ell^2}} \label{eq:dgp_kernel}\\
    &\approx w^2 \sigma_{ij} e^{-\frac{Q_{x, x^\prime}}{2}} = k_{NS}(x, x^\prime|\Sigma_i),\\
    \Sigma_i &= \frac{2\ell^2}{f_{l-1}'^2(x_i)}.\label{eq:dgp_lens}
\end{align}
\begin{figure*}[t]
    \centering
    \includegraphics[width=\textwidth]{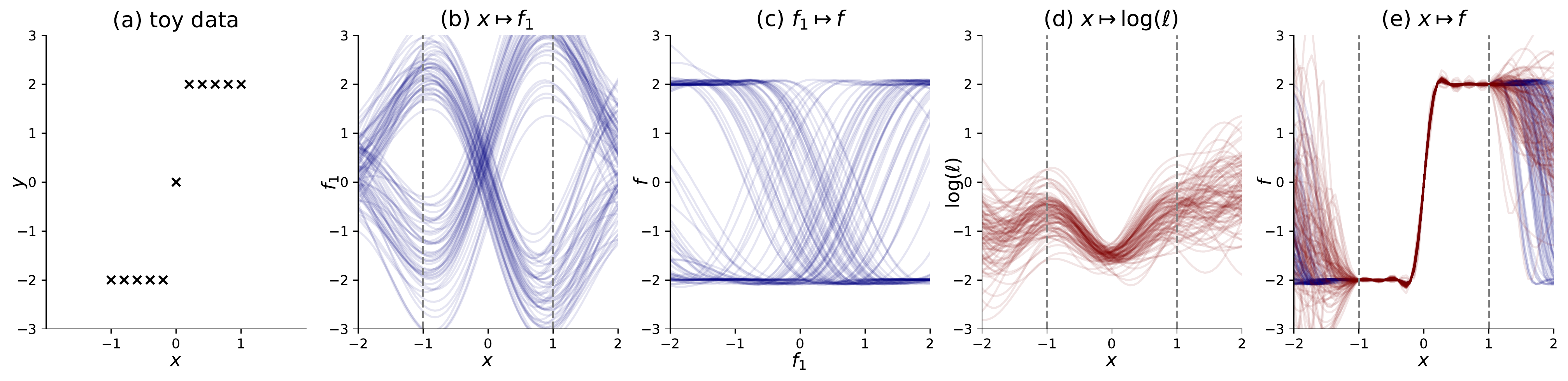}\\
    \caption{Posterior HMC draws on two DGPs. \textbf{(b)} and \textbf{(c)} demonstrate functions that are translations or mirror images of each other, while the multi-modal behavior disappears in \textbf{(d)}.}
    \label{fig:toy}
\end{figure*}
\hspace{-0.11cm}Here the $Q_{x, x^\prime}$ corresponds to the squared distance defined in \eqref{eq:qij}. According to our analysis (Table \ref{tab:spectrograms}), the DGP kernel \eqref{eq:dgp_kernel} and an NSQ kernel \eqref{eq:nsq} with a lengthscale parametrization \eqref{eq:dgp_lens} share the same spectrogram, demonstrating an equivalence up to second-order effects.\par
\subsection{Advantages of covariance function DGPs}
Given the equivalence drawn in \ref{sec:dgp_equiv}, we delineate the pros and cons between compositional DGPs and DGPs formulated with covariance functions \citep{dunlop2018deep}, which comprise a layered structure of nonparametric kernels $k_{\bt}$:
\begin{align}
    f_0(x) &\sim \GP(0, k_{\bt}(x, x'| \bt=\bt_0)),\\
    f_l(x) | f_{l-1}(x) &\sim \GP(0, k_{\bt}(x, x'|\bt=F\circ f_{l-1})). \label{eq:dgp_cov}
\end{align}
Here $k_{\bt}$ denotes such nonparametric kernels as NSQ \eqref{eq:nsq} \citep{paciorek2004nonstationary}, $\bt$ their functional hyperparameters, and $F$ a warping function \eqref{eq:warped} mapping GP samples to valid parameters. Despite the notable equivalence, the two types of GP models behave differently in practice \citep{dunlop2018deep}. This discrepancy warrants a closer investigation, which evidences the superiority of covariance function DGPs.\par
We posit that the deep compositional probabilistic models \eqref{eq:dgp_def} could benefit from monotonicity in hidden layers $f_0, f_1 \cdots, f_{L-1}$, a constraint not required for covariance function DGPs \eqref{eq:dgp_cov}. Zero-mean compostional DGPs exhibit a pathology where the prior space of $f_{L}\circ\cdots\circ f_0$ degenerates into piecewise constant functions as $L \rightarrow \infty$ \citep{duvenaud2014avoiding, dunlop2018deep}. The state-of-the-art DGPs \citep{salimbeni2017doubly, havasi2018inference} remedy this pathology with calibrated mean functions, which are likely to generate monotonic sample paths, despite the seemingly detrimental effect on model expressivity. Our derivation of approximate equivalence holds when the function $f_{l-1}(x)$ has nonzero derivatives almost everywhere, and is consequently monotonic. The ``equivalence conditioned on monotonicity'' marks the absence of rank pathology in covariance function DGPs \eqref{eq:dgp_cov} and provides alternate justification on calibrated mean functions.\par
Covariance function DGP avoids multi-modality by directly modeling lengthscale values. 
\begin{figure*}[t]
    \centering
    \includegraphics[width=0.8\textwidth]{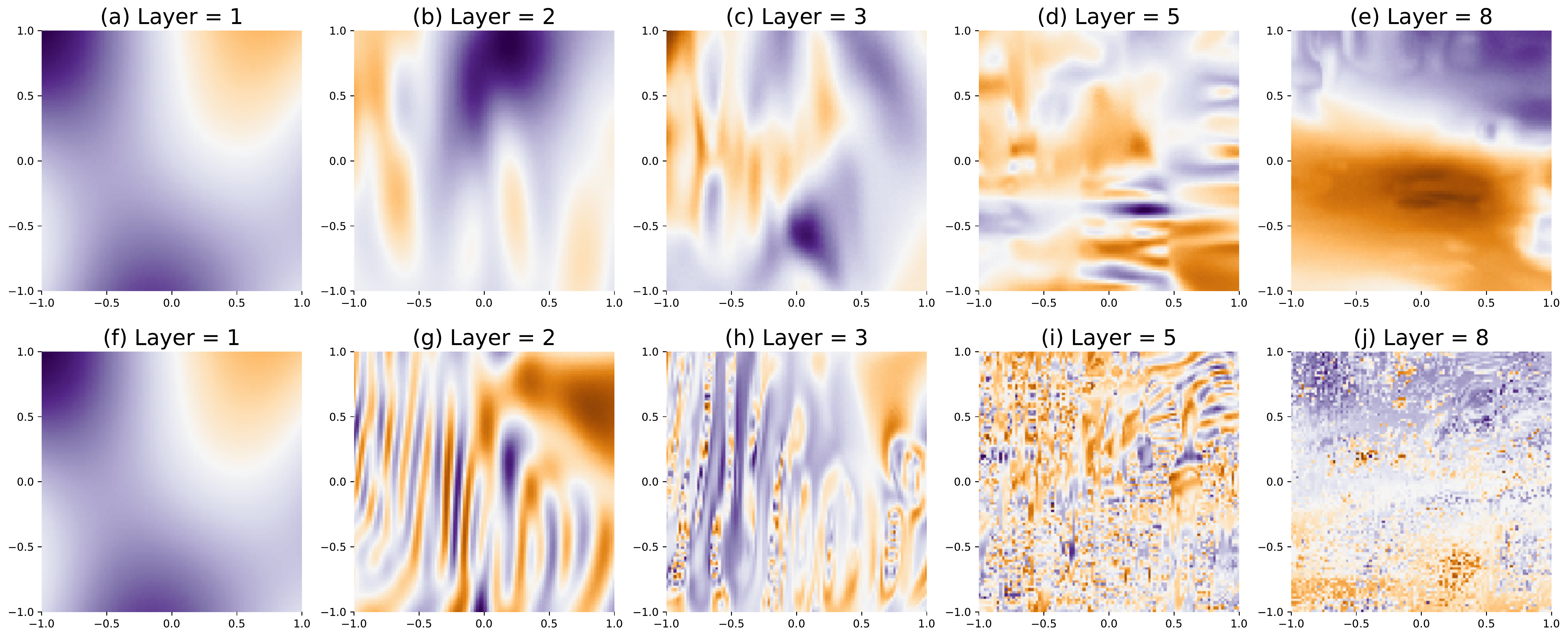}\\
    \caption{2D prior draws from covariance function DGPs with correlation function NSQ (first row) and CSK \eqref{eq:csk_norm} (second row).}
    \label{fig:deep}
\end{figure*}
\begin{figure*}[b]
    \centering
    \includegraphics[width=0.85\textwidth]{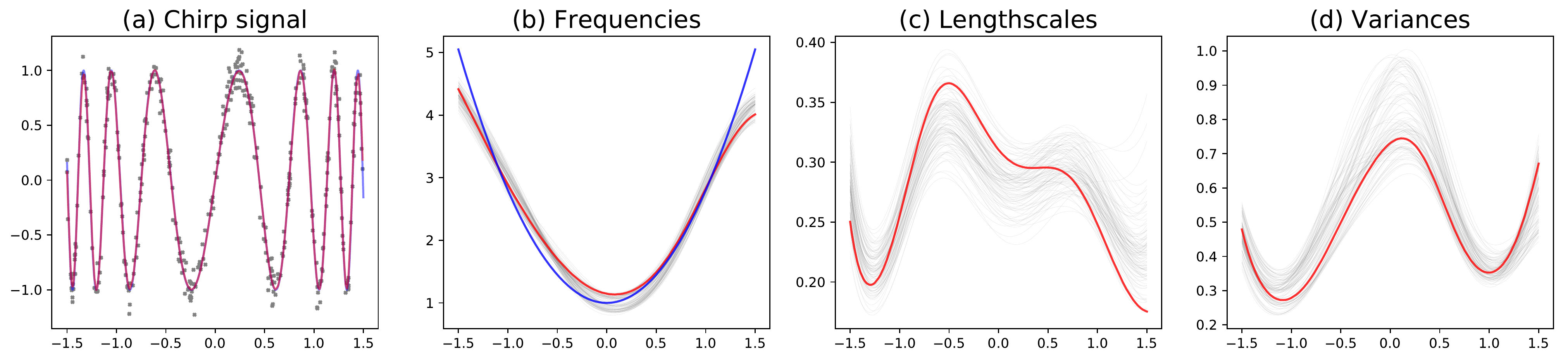}
    \caption{Regression on synthetic chirp signal. \textbf{(a)}: The training data (grey points) with the ground truth line (blue) and predicted mean (red); ground truth and predict mean overlap. \textbf{(b)}: HMC samples of the learned frequency function, with ground truth instantaneous frequency (blue line) and MAP estimate (red line). \textbf{(c)}, \textbf{(d)}: HMC samples of lengthscale and variance functions, with MAP estimates marked with red lines.}
    \label{fig:chirp}
\end{figure*}
It is worth noting that the conditional kernel \eqref{eq:dgp_kernel} stays invariant under \emph{translation} and \emph{reflection} of $f_{l-1}$, thus defining an equivalence class containing the above transformations. The invariance yields highly multi-modal posterior distributions \citep{havasi2018inference}, a major obstacle to effective inference. The linear mean function \citep{salimbeni2017doubly} solves reflection but not translation of of $f_{l-1}$. Meanwhile, functions belonging to the same equivalence class correspond to a singular lengthscale \eqref{eq:dgp_lens}, and consequently a single covariance function DGP, which has notably more concentrated posterior densities. We demonstrate the two posteriors of a toy example in Figure \ref{fig:toy}.\par
While the covariance function DGP appears to claim a minor turf \citep{paciorek2004nonstationary, heinonen2016non, remes2017non} in deep probabilistic modeling, we advocate that it replace the customary compositional DGPs as an equivalent but more effective alternative.

\subsection{CSK as deep Gaussian processes}
Applying CSK \eqref{eq:csk_final} in the covariance function DGP recursion \eqref{eq:dgp_cov} yields a DGP that nests the NSQ construction \citep{dunlop2018deep} as a special case, defining a proven \emph{ergodic} Markov chain \citep{dunlop2018deep}. While the ergodicity of the DGP-CSK model upper bounds the model complexity with its mixing time, the application of CSK, nevertheless, significantly enriches the prior space (Figure \ref{fig:deep}).

\section{Scalable inference for covariance function DGPs}
In this section, we discuss the understudied scalable inference schemes for covariance function DGPs \citep{dunlop2018deep}. Our inference framework places inducing points on functional kernel parameters $\bt$ and the function $f$. It shares the same time complexity as inducing point sparse DGPs \citep{salimbeni2017doubly, havasi2018inference}. Consider a nonparametric kernel $k_{\bt}(\cdot, \cdot)$, where the functional hyperparameters $\bt$ have warped GP priors \eqref{eq:warped}. Possible candidates of kernels include the NSQ kernel \eqref{eq:nsq}, GSM kernel \eqref{eq:gsm} and CSK \eqref{eq:csk_final}. Without loss of generality, we assume a 2-layer deep GP ($L=1$):
\begin{align}
    f &\sim \GP(0, k_{\bt}(\cdot,\cdot)),\\
    \theta_d(\x) &= F_d(h_d(\x)), h_d(\x) \sim \GP(0, k_d(\cdot, \cdot)).
\end{align}
With noisy observations $\{\X, \y\}$ \eqref{eq:obs}, inducing points $\z_f, \z_\theta$ and their respective observations $\u_f, \u_{\bt}$, we can factorize the joint likelihood:
\begin{align}
    p(\y, \f, \bt, \u) &= \underbrace{p(\y|\f)}_{\text{likelihood}}\underbrace{p(\f|\u_f, \bt)p(\u_f|\bt)p(\bt|\u_{\bt})p(\u_{\bt})}_{\text{prior}}, \label{eq:sparse_lik}
\end{align}
where $\bt=\bt\left(\X\bigcup\z_f\right)$. The posterior distribution $p(\u_{\bt}, \u_f|\y)$ is intractable, and we approximate it with stochastic gradient Hamiltonian Monte Carlo (SG-HMC) \citep{chen2014stochastic, havasi2018inference} and doubly stochastic variational inference (DS-VI) \citep{salimbeni2017doubly}. While we use 2-layer DGP as an example, the likelihood factorizes analogously for more layers.
\subsection{Stochastic gradient Hamiltonian Monte Carlo}
Hamiltonian Monte Carlo (HMC) \citep{Neal93probabilisticinference} produces samples from the posterior distribution $p(\u|\y)$, which requires the gradient computation for the negative log-posterior $U(\u) = -\log p(\u|\y)$. SG-HMC \citep{chen2014stochastic} scales up classical HMC with stochastic gradient estimates.\par
Using Jensen's inequality, we can observe from \eqref{eq:sparse_lik} that the lower bound of the joint log-likelihood $p(\u_{\bt}, \u_f, \y)$ can be approximated via Monte Carlo sampling:
\begin{align}
    \log p(\u_{\bt}, \u_f, \y) 
    &\geq \mathbb{E}_{p(\f|\u_f, \bt)p(\bt|\u_{\bt})} \log \frac{p(\y, \f, \bt, \u)}{p(\f|\u_f, \bt)p(\bt|\u_{\bt})} \notag\\
    &\approx \log p(\y|\widehat{\f})p(\u_f|\widehat{\bt})p(\u_{\bt}). \label{eq:sghmc}
\end{align}
The vectors $\widehat{\bt}, \widehat{\f}$ are Monte Carlo samples: $\widehat{\bt}\sim p(\bt|\u_{\bt}), \widehat{\f} \sim p(\f|\u_f, \widehat{\bt})$. We can thus approximate the gradient using \eqref{eq:sghmc}, where
\begin{align}
\nabla U(\u_{\bt}, \u_f)=-\nabla\log p(\u_{\bt}, \u_f|\y) = -\nabla \log p(\u_{\bt}, \u_f, \y).
\end{align}
We refer to Section 3 in the appendix for hyperparameter training and stochastic variational inference for this model.\par
\section{Experiments}
\begin{figure*}[t]
    \centering
    \includegraphics[width=0.9\textwidth]{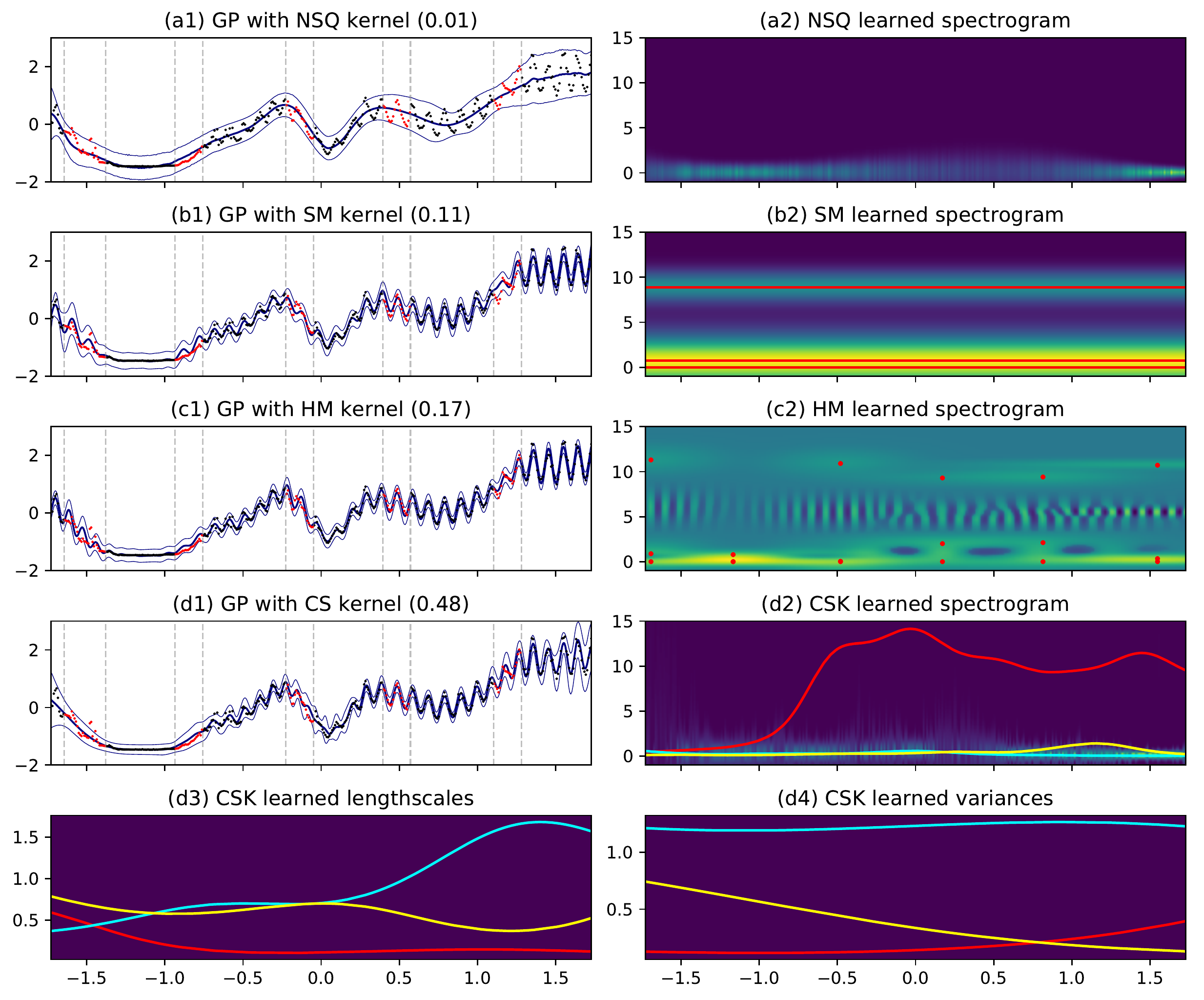}
    \caption{GP regression with solar irradiance. Test log-likelihoods are shown in parentheses for figures \textbf{(a1)}-\textbf{(d1)}. The learned frequencies are marked by the red lines and points on the spectrogram plot \textbf{(b2)}-\textbf{(d2)}. The lengthscale and variance function learned from CSK is plotted on \textbf{(d3)}-\textbf{(d4)}.}
    \label{fig:solar}
\end{figure*}

\begin{figure*}[t]
    \centering
    \includegraphics[width=\textwidth]{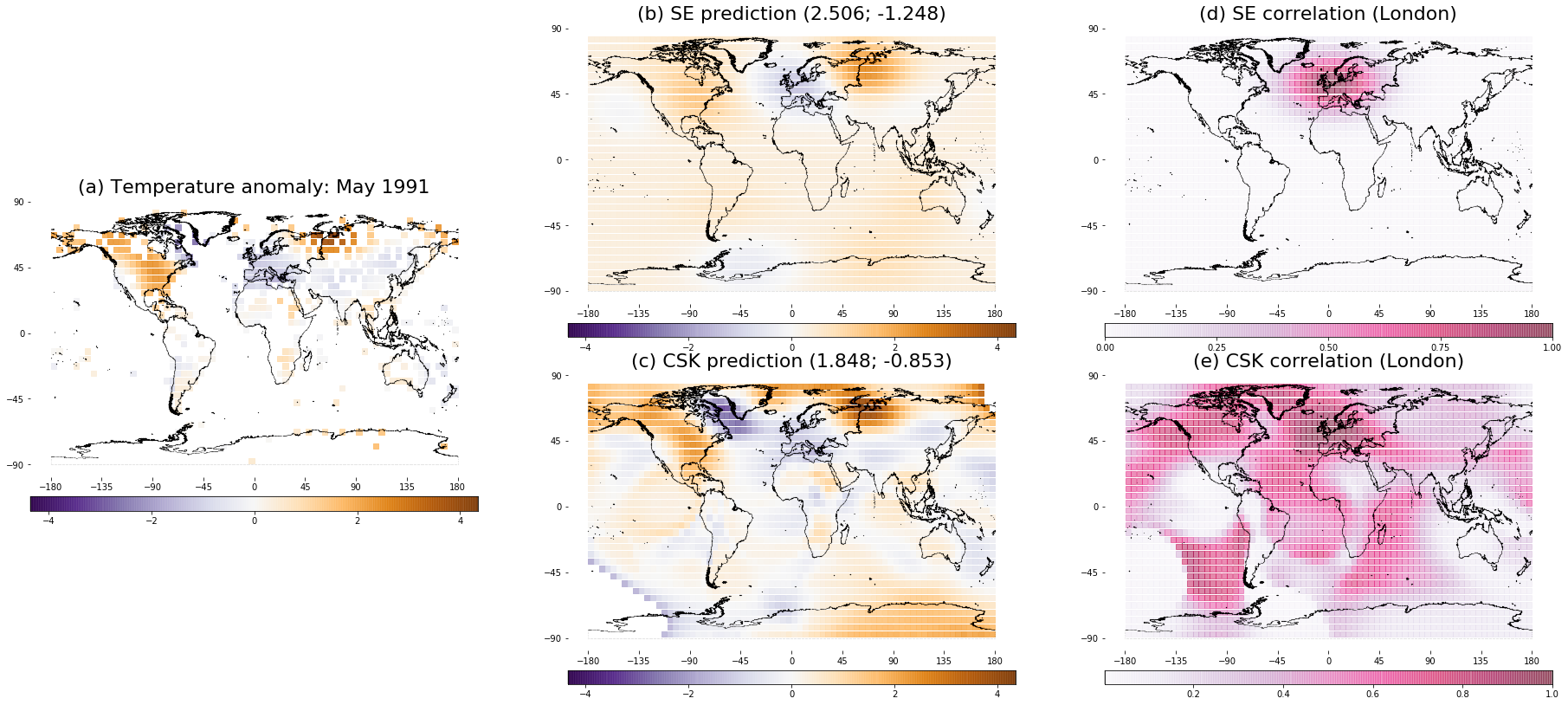}
    \caption{Air temperature anomaly dataset. \textbf{(a)} demonstrates the temperature anomaly readings from May 1991. \textbf{(b)}, \textbf{(c)} display the posterior predictive mean for May 1991 on a grid of global locations, with the numbers in parentheses denoting mean squared error (MSE) and mean log-likelihood, respectively. \textbf{(d)}, \textbf{(e)} depict the correlation between London and other geographical locations: it is worth noting that the SE kernel \textbf{(d)} only captures positive correlation on a small elliptical region.}
    \label{fig:air}
\end{figure*}

\subsection{Recovering chirp signal}
We first tested our methods on a simulated dataset. A chirp signal is a non-stationary signal taking the generic form $x(t)=\cos(\phi(t))$. In this setting, we take 400 noisy observations of a chirp signal $x(t) = \cos(2\pi(t+0.6t^3))$, and train a one-component CSK with 30 inducing points. The synthetic instantaneous frequency of this signal is $\phi'(t)/2\pi=1+1.8t^2$, which is recovered with the frequency term in CSK (see figure \ref{fig:chirp}).
\subsection{Solar irradiance}
We consider regression on the solar irradiance dataset \citep{lean2004solar}, which has shown some non-stationarities. We compared GP models with NSQ, SM, HMK \citep{shen2019harmonizable} and CSK, where the inference for SM and HMK were done with sparse GP regression with inducing points \citep{titsias2009variational}, and the functional hyperparameters of NSQ and CSK are inferred with SGHMC as illustrated in the previous section. While it is not immediately clear from the spectrogram visualizations (see figure \ref{fig:solar}, the spectral kernels (SM, HMK and CSK) learns similar frequency patterns: SM learns global frequency peaks exhibited in data; HMK learns and interpolates local patterns; CSK learns a global pattern with varying frequencies, while also accounting for the non-Gaussianity of the data.
\vspace*{-0.2cm}
\subsection{Air temperature anomaly dataset}
We conducted spatiotemporal analysis on the air temperature anomaly dataset \citep{jones1994hemispheric}, which contains monthly air temperature deviations from the monthly mean temperature measured on locations on a global grid. We subsampled the data from 1988-1993 with $32910$ readings and partitioned an 80\%-20\% split on training and testing data. Figure \ref{fig:air} demonstrates the predictive temperature anomaly between a GP with SE kernel and CSK with 5 components, which significantly improves the predictive performance. Nonstationarity is required to capture the correlation patterns demonstrated in the data. 
\begin{table*}[t]
    \centering
    \resizebox{\textwidth}{!}{
    \begin{tabular}{l|ccccccc}\toprule
        Model & LINEAR & VANILLA 100 & VANILLA 500 & SM & DGP 4 & NSQ & CSK\\
        \hline
        Test log-likelihood & -1.278 & -0.847 & -0.823 & -0.805 & -0.623 & -0.580 & -0.576\\
        Test MSE & 0.755 & 0.333 & 0.314 & 0.298 & 0.303 & 0.293 & 0.289\\
        \bottomrule
    \end{tabular}
    }
    \caption{GP regression results for NY taxi dataset.}
    \vspace{-0.5cm}
    \label{tab:taxi}
\end{table*}
\subsection{New York Yellow Taxi dataset}
We ran GP regression on a subset of the New York Yellow Taxi dataset\footnote{\url{http://www.nyc.gov/html/tlc/html/about/trip_record_data.shtml}}, whose objective is to predict the taxi trip duration given the pickup and dropoff locations and the starting date and time. Given CSK's ability to handle periodicities, we treat the date and time as one feature. We ran 7 different models in total: Bayesian linear regression (LINEAR), vanilla sparse GPs with 100 and 500 inducing points (VANILLA 100 \& 500), sparse GP with SM kernel (SM), 4-layer compositional deep GP with ``monotonic'' inner layers (DGP 4), GP with NSQ and CSK. Apart from VANILLA 500, other models were run with 100 inducing points.\par
One can tell from comparison in Table \ref{tab:taxi} that the taxi dataset is nonlinear, non-Gaussian and exhibits nontrivial frequency patterns. CSK marginally outperforms better than other model by accounting for all three properties. NSQ outperforms a 4-layer DGP due to a different parametrization: NSQ is roughly equivalent to a 2-layer DGP where the inner layers for each data dimensions are modelled with SE kernels with independent lengthscales.

\section{Conclusion}
In this work, we propose the \emph{convolutional spectral kernel}, which generalizes \citet{paciorek2004nonstationary} with spatially varying frequencies. We analyze common kernels and GP models having input warping with spectrogram, which sheds light on the interpretation of deep models, and draws an equivalence between two types of DGPs. We propose a novel scalable inference framework for DGPs constructed via covariance functions, which empirically outperforms current compositional DGP methods. The theoretical results derived in this paper indicate that covariance function DGPs, as an appealing alternative to current DGPs, warrant further study.
\bibliographystyle{plainnat}
\bibliography{refs}

\newpage
\onecolumn
\section*{Supplementary materials}
\section*{1\ \ Convolutional spectral kernels: a derivation}
In this section, we derive the convolutional spectral kernel (CSK) in detail, which solves the following integral:
\begin{align}
    K_{\x_i}(\x_i-\u) &= \N(\x_i-\u|\imath\bmu_i, \bS_i)=\frac{1}{|2\pi\bS_i|^{1/2}}\exp\left(-\frac{1}{2}(\x_i-\u-\imath\bmu_i)^\top\bS^{-1}(\x_i-\u-\imath\bmu_i)\right),\\
    k(\x_i, \x_j) &= \int K_{\x_i}(\x_i-\u)\overline{K_{\x_j}(\x_j-\u)}\d{\u}.
\end{align}
Note that a transpose instead of Hermitian is used in the feature representation, making the function $K_{\x_i}(\x_i-\u)$ an improper density. We denote the Fourier transform as an operator on functions: $\mathcal{F}\left[f\right](\bomega) = \int f(\x)e^{\imath\bomega^\top\x} \d{\x}$. The Fourier transform of $K_{\x_i}(\u)$ takes a simple form, which can be used to formulate the Fourier transform of the kernel:
\begin{align}
    \mathcal{F}\left[\N\left(\mathbf{v}|\imath\bmu, \bS\right)\right] &= \exp\left(-\frac{1}{2}\bomega^\top\bS\bomega-\bmu^\top\bomega\right),\\
    k(\x_i-\x_j) &= \mathcal{F}^{-1}\left\{\mathcal{F} \left[k(\x_i-\x_j)\right]\right\}\\
     &= \mathcal{F}^{-1}\left\{\mathcal{F} \left[\int K_{\x_i}(\x_i-\u)\overline{K_{\x_j}(\x_j-\u)}\d{\u}\right]\right\}\\
    &= \mathcal{F}^{-1}\left\{\mathcal{F} \left[K_{\x_i}(\x_i-\u)\right]\mathcal{F} \left[\overline{K_{\x_j}(\x_j-\u)}\right]\right\}\\
    &= \mathcal{F}^{-1}\left\{\mathcal{F} \left[K_{\x_i}(\x_i-\u)\right]\overline{\mathcal{F}} \left[K_{\x_j}(\x_j-\u)\right]\right\}\\
    &= \mathcal{F}^{-1}\left\{\mathcal{F} \left[K_{\x_i}(\x_i-\u)\right]\mathcal{F} \left[K_{\x_j}(\x_j-\u)\right]\right\}\\
    &= \mathcal{F}^{-1}\left\{\exp\left(-\frac{1}{2} \bomega^\top\bS_i\bomega-\bmu_i^\top\bomega-\frac{1}{2} \bomega^\top\bS_j\bomega-\bmu_j^\top\bomega\right)\right\}\\
    &= \mathcal{F}^{-1}\left\{\exp\left(-\frac{1}{2} \bomega^\top(\bS_i+\bS_j) \bomega - (\bmu_i+\bmu_j)^\top\bomega\right)\right\}\\
    &= \N\left(\x_i-\x_j|\imath(\bmu_i+\bmu_j), \bS_i+\bS_j\right).
\end{align}
The pseudo-Gaussian density gives a convenient closed-form Fourier transform, which we use to obtain the solution to the Hermitian inner product.
\section*{2\ \ The spectrogram derivations}
Many kernels share the form of a mixture of generalized Gaussian characteristic functions (CFs). In this section, we address certain qualities of the Gaussian characteristic functions that justify our approximation of the kernel. Consider the CF of a Gaussian distribution (spectral mixture kernel \citep{wilson2013gaussian}) the following generalized Gaussian CF:
\begin{align}
    k_{SM}(\x_i, \x_j) &= \sigma\exp\left(-\frac{1}{2}(\x_i-\x_j)^\top\bS^{-1}(\x_i-\x_j)+\imath\langle\bmu, \x_i-\x_j\rangle\right), \label{eq:supp_sm}\\
    k(\x_i, \x_j) &=  \sigma_{ij}\exp\left(-\frac{1}{2}D_{ij}+\imath U_{ij}\right). \label{eq:supp_sm_generalized}
\end{align}
The Wigner transform is tractable for the SM kernel \eqref{eq:supp_sm}, because $D_{ij}$ and $U_{ij}$ terms are linearized with respect to $\btau = \x_i-\x_j=(\x+\btau/2)-(\x-\btau/2)$. We can similarly linearize the two terms for an approximation of kernel values, and get an approximate Wigner transform. Given by the property of a Gaussian CF, we give the following assumptions based on the convexity of $D_{ij}$ and the zero diagonal of $U_{ii}$ for $k_{SM}$.
\begin{align}
    D_{\x, \x} &= 0, \\
    \mathcal{J}_{\t}D_{\x+\t/2,\x-\t/2} &= \mathbf{0},\\
    \mathcal{H}_{\t} D_{\x+\t/2,\x-\t/2} &\succeq \mathbf{0},\\
    U_{\x,\x} &= 0.
\end{align}
We approximate a Taylor expansion based on those assumptions:
\begin{align}
   D_{\x+\btau/2, \x-\btau/2} &\approx D_{x, x} + \langle\left.\mathcal{J}_{\t}D_{\x+\t/2,\x-\t/2}\right\vert_{\t=\mathbf{0}}, \btau\rangle + \btau^\top \left.\mathcal{H}_{\t} D_{\x+\t/2,\x-\t/2}\right\vert_{\t=\mathbf{0}}\btau\\
    &= \btau^\top \left.\mathcal{H}_{\t} D_{\x+\t/2,\x-\t/2}\right\vert_{\t=\mathbf{0}}\btau\\
    &= \btau^\top\bL_{\x}\btau,\\
    U_{\x+\btau/2, \x-\btau/2} &\approx U_{\x, \x} + \langle\left.\mathcal{J}_{\t}U_{\x+\t/2,\x-\t/2}\right\vert_{\t=\mathbf{0}}, \btau\rangle\\
    &= \langle\left.\mathcal{J}_{\t}U_{\x+\t/2,\x-\t/2}\right\vert_{\t=\mathbf{0}}, \btau\rangle\\
    &= \langle\boldsymbol{\xi}_{\x},\btau\rangle.
\end{align}
The linearization gives an approximate of kernel values where $\x$ and $\btau$ are separate:
\begin{align}
    k(\x+\btau/2, \x-\btau/2) &\approx \sigma_{\x\x}\exp\left(-\frac{1}{2}\btau^\top\bL_{\x}\btau+\imath \langle\boldsymbol{\xi}_\x, \btau\rangle\right)\\
    &=\widehat{k}(\x+\btau/2, \x-\btau/2),\\
    W(\x, \bomega) &= \int k(\x+\btau/2, \x-\btau/2)e^{-2\imath\pi\langle\bomega, \btau\rangle}\d{\btau}\\
    &\approx \int \widehat{k}(\x+\btau/2, \x-\btau/2)e^{-2\imath\pi\langle\bomega, \btau\rangle}\d{\btau}\\
    &\approx \sigma_{\x\x}\N\left(\left.\bomega\right\vert \frac{\boldsymbol{\xi}_{\x}}{2\pi}, \frac{\bL_{\x}}{2\pi^2}\right)\\
    &= \widehat{W}(\x, \bomega),
\end{align}
which gives the spectrogram as the approximate Wigner distribution function. The rest of this subsection derives the corresponding $\boldsymbol{\xi}_\x$ and $\bL_{\x}$ for some GP models.
\subsection*{2.1 \ \ The non-stationary quadratic (NSQ) kernel}
The NSQ kernel \citep{paciorek2004nonstationary} is a special case of CSK:
\begin{align}
    k_{NS}(\x_i, \x_j) &= \frac{|\bS_i|^{1/4}|\bS_j|^{1/4}}{|(\bS_i+\bS_j)/2|^{1/2}}e^{-\frac{D_{ij}}{2}}, \label{eq:supp_nsq}\\
    D_{ij} &= (\x_i-\x_j)^\top\left(\bS_i+\bS_j\right)^{-1}(\x_i-\x_j), \bS_i\succeq \mathbf{0},\\
    U_{ij} &\equiv 0.
\end{align}
It is straightforward that $\boldsymbol{\xi}_\x = \mathbf{0}$. Now we derive $\bL_{\x}$:
\begin{align}
    \mathcal{H}_{\t} D_{\x+\t/2,\x-\t/2} &= \mathcal{H}_{\t} \t^\top\left(\bS_{\x+\t/2}+\bS_{\x-\t/2}\right)^{-1}\t\\
    &= (\bS_{\x+\t/2} + \bS_{\x-\t/2})^{-1}.
\end{align}
Other terms involving the Hessian operator are $\mathbf{0}$ because of our assumptions. When $\t=\mathbf{0}$, we get $\bL_{\x}=\frac{1}{2}\bS_{\x}^{-1}$.
\subsection*{2.2 \ \ The generalized spectral mixture (GSM) kernel}
The GSM kernel \citep{remes2017non} augments the NSQ kernel with a cosine term:
\begin{align}
    k_{GSM}(\x_i, \x_j) &= k_{NS}(\x_i, \x_j) \exp\left(\imath U_{ij}\right),\\
    U_{ij} &= \langle\bmu_i, \x_i\rangle - \langle\bmu_j, \x_j\rangle.
\end{align}
The lengthscale function conforms to the NSQ kernel: $\bL_\x=\frac{1}{2}\bS_{\x}^{-1}$. The frequencies are derived as:
\begin{align}
    \left.\mathcal{J}_{\t} U_{\x+\t/2, \x-\t/2}\right\vert_{\t=\mathbf{0}} &= \mathcal{J}_{\t} \left[\bmu_{\x+\t/2}^\top(\x+\t/2)-\bmu_{\x-\t/2}^\top(\x-\t/2)\right]\\
    &=\left.\mathcal{J}_{\t} \left[\left(\bmu_{\x+\t/2}-\bmu_{\x-\t/2}\right)^\top\x + \left(\frac{\bmu_{\x+\t/2}+\bmu_{\x-\t/2}}{2}\right)^\top\t\right]\right\vert_{\t=\mathbf{0}}\\
    &= \bmu_{\x} + \left(\mathcal{J}_{\x}\bmu_{\x}\right)\x.
\end{align}
\subsection*{2.3 \ \ The Deep Gaussian process (DGP)}
The kernel of DGP is a SE kernel with input warping $\f_i = \f(\x_i)$. Formally,
\begin{align}
    k(\x_i, \x_j) &= \exp\left(-\frac{1}{2}(\f_i-\f_j)^\top\bL(\f_i-\f_j)\right),\\
    D_{ij} &= (\f_i-\f_j)^\top\bL(\f_i-\f_j).
\end{align}
The lengthscale function differ from a constant $\bL$ because $\f_i$ is a function of $\x_i$.
\begin{align}
    \bL_{\x} &= \left(\mathcal{J}_{\x} \f\right)^\top \bL\left(\mathcal{J}_{\x} \f\right).
\end{align}

\section*{3 \ \ Inference with covariance function DGPs}
\subsection*{3.1\ \ Previous work: maximum \emph{a-posteriori} and Hamiltonian Monte Carlo}
Previous work \citep{paciorek2004nonstationary, heinonen2016non, remes2017non} have studied inference for the functional hyperparameters of nonparametric kernels. A maximum \emph{a-posteriori} framework gets MAP estimates for values of the hyperparameters on the input points:
\begin{align}
    p(\bt | \X) \propto p(\X, \bt) = p(\y|\f) p(\f | \bt)p(\bt), \label{eq:map}
\end{align}
Maximizing the posterior likelihood \eqref{eq:map}, we obtain point estimates for $\bt(\X)$.\par
One point estimate with maximum likelihood does not effectively represent the posterior distribution, the Hamiltonian Monte Carlo (HMC) \citep{Neal93probabilisticinference, heinonen2016non} produces samples from $p(\boldsymbol{\Theta} | \X)$ using Hamiltonian dynamics. MAP and HMC are solid choices for small-scale data, but is infeasible in large-scale setting due to the necessity of inferring $O(N)$ parameters.
\subsection*{3.2\ \ Monte Carlo expectation maximization and the SG-HMC sampler}
For hyperparameter learning using SG-HMC, we use the moving window Monte Carlo expectation maximization (MCEM) \citep{havasi2018inference}, an extension of the expectation maximization algorithm to learn maximum likelihood estimate of hyperparameters. The moving window MCEM keeps track of a fixed-length window of recent samples from HMC sampler, and each optimization action of hyperparameters is done given the parameters randomly drawn from the window. This algorithm has shown better optimization of hyperparameter for deep GP models \citep{havasi2018inference}.\par
In practice, we use the auto-tuning approach given by \citep{springenberg2016bayesian}, which has shown to work well for Bayesian neural networks, as well as compositional DGPs \citep{havasi2018inference}.
\subsection*{3.3 \ \ Doubly stochastic variational inference: a derivation}
\citet{salimbenideeply} proposes a DS-VI framework for DGPs with NSQ covariance with omitted derivation. Here we provide a detailed derivation. The main idea is to approximate the evidence lower bound with Monte Carlo samples. We first consider variational distributions that factorize over $f$ and $\bt$: $q\left(\u_f, \u_\bt\right) = \N(\u_f|\mathbf{m}_f,
\mathbf{S}_f)\N(\u_\bt|\mathbf{m}_\bt, \mathbf{S}_\bt)$. we can thus get the approximated posterior:
\begin{align}
    q(\f, \bt, \u_f, \u_\bt) &= p(\f|\u_f, \bt)q(\u_f) p(\bt|\u_\bt)q(\u_\bt).
\end{align}


Here we can employ a Monte Carlo sampling of $\widehat{\bt}$. A stochastic estimate of ELBO can be obtained:
\begin{align}
    \mathcal{L}_{\text{CSK}} &= \log p(\y|\widehat{\f}) - \text{KL}(q(\u_f)\|p(\u_f | \widehat{\bt})) - \text{KL}(q_(\u_\bt)|p(\bt)).
\end{align}
Empirically, we use a whitened representation of the variational parameter: $\check{\u}_\theta = k_\theta(\Z_\theta, \Z_\theta)^{-1/2}\u_\theta$.\par
\section*{4\ \ Experiment details}
Our experiments mainly include the following models: the sparse Gaussian process regression \citep{titsias2009variational}, the sparse variational Gaussian process \citep{hensman13gaussian}, the doubly stochastic variational deep Gaussian process \citep{salimbeni2017doubly} and the proposed stochastic gradient Hamiltonian Monte Carlo method for inferring covariance function DGPs. 
\subsection*{4.1\ \ Solar irradiance}
With the solar irradiance dataset \citep{lean2004solar}\footnote{\url{https://github.com/jameshensman/VFF/blob/master/experiments/solar/solar_data.txt}}, we follow the same partition of training and test set as \citet{gal2015improving} and \citet{hensman2017variational}. This 1-dimensional dataset has 281 training points and 110 test points. We standardize both $\X$ and $\Y$ before the regression.\par
We use 3 components to fit the SM kernel as well as CSK. The HMK kernel has 6 centroids, each with three frequency terms. We initialize all parameters (including kernel hyperparameters for parametric kernels, kernel hyperparameters for latent GPs of NSQ and CSK, variational parameters, inducing points, and parameters sampled from HMC) at random. And we do 5000 iterations for each models, with a learning rate of $10^{-3}$.
\subsection*{4.2\ \ New York Yellow Taxi dataset}
Due to limited time and CSK's sensitivity to higher dimensions, we use a subset of the New York Taxi dataset\footnote{\url{https://www.kaggle.com/c/nyc-taxi-trip-duration/overview}}, subsampling 25\% of all taxi trips from the first two months (Jan-Feb 2016). The dataset has 5 dimensions (longitudes and latitudes of the pickup and dropoff locations, and the date and time of the beginning of the taxi trip). We combine the date and time to a ``date value''. The training set includes 100671 data points, and the test set includes 45260 data points. We also standardize both input and output before regression.\par
We use minibatch of 10000 given the medium-large scale of data. We use the stochastic variational GP \citep{hensman13gaussian} for GPs with parametric kernels (SE kernel and SM kernel), and SG-HMC \citep{havasi2018inference} for compositional DGP, and the proposed SG-HMC for GP with NSQ kernel and CSK. For CSK and SM, we use 4 frequency components. The parameters are again initialized at random, except for the inducing points randomly initialized with a K-Means algorithm. 
\end{document}